\documentclass[conference]{IEEEtran}
\IEEEoverridecommandlockouts

\usepackage{cite}
\usepackage{url}
\usepackage[utf8]{inputenc}
\usepackage[T1]{fontenc}    % use 8-bit T1 fonts
\usepackage{amsmath,amssymb,amsfonts}
\usepackage{graphicx}
\usepackage{textcomp}
\usepackage{xcolor}
\usepackage{etoolbox} 

\usepackage{booktabs}       % professional-quality tables
\usepackage{nicefrac}       % compact symbols for 1/2, etc.
\usepackage{multirow} 
\usepackage{microtype}      % microtypography
\usepackage[detect-weight=true,detect-inline-weight=math]{siunitx}
\usepackage[capitalise]{cleveref}
\usepackage{makecell}
\usepackage{longtable}

\def\real{\mathbb R}

\def\gd{\text d}
\def\est{\widehat} 
\def\comment#1{}

\def\ones{{\bf1}}
\def\diag#1{\text{diag}\left(#1\right)}
\def\r#1{\left[#1\right]}

\def\eqref#1{(\ref{#1})}
\def\Beq#1\Eeq{\begin{equation}#1\end{equation}}
\def\Beqo#1\Eeqo{\begin{equation*}#1\end{equation*}}
\def\Beqs#1\Eeqs{\begin{align}#1\end{align}}
\def\Beqso#1\Eeqso{\begin{align*}#1\end{align*}}

\begin{document}

\title{Least Redundant Gated Recurrent Neural Network}

\author{\IEEEauthorblockN{1\textsuperscript{st} \L{}ukasz Neumann}
\IEEEauthorblockA{\textit{Institute of Computer Science} \\
\textit{Warsaw University of Technology}\\
Warsaw, Poland \\
lukasz.neumann@pw.edu.pl}
\and
\IEEEauthorblockN{2\textsuperscript{nd} \L{}ukasz Lepak}
\IEEEauthorblockA{\textit{Institute of Computer Science} \\
\textit{Warsaw University of Technology}\\
Warsaw, Poland \\
lukasz.lepak.dokt@pw.edu.pl}
\and
\IEEEauthorblockN{3\textsuperscript{rd} Pawe\l{} Wawrzy\'nski}
\IEEEauthorblockA{\textit{Ideas NCBR}\\
Warsaw, Poland \\
pawel.wawrzynski@ideas-ncbr.pl}
}

\maketitle

\begin{abstract}
Recurrent neural networks are important tools for sequential data processing. However, they are notorious for problems regarding their training. Challenges include capturing complex relations between consecutive states and stability and efficiency of training. In this paper, we introduce a~recurrent neural architecture called Deep Memory Update (DMU). It is based on updating the previous memory state with a~deep transformation of the lagged state and the network input. The architecture is able to learn to transform its internal state using any nonlinear function. Its training is stable and fast due to relating its learning rate to the size of the module. Even though DMU is based on standard components, experimental results presented here confirm that it can compete with and often outperform state-of-the-art architectures such as Long Short-Term Memory, Gated Recurrent Units, and Recurrent Highway Networks.
\end{abstract}


\section{Introduction} 

Recurrent Neural Networks (Recurrent NNs, RNNs) are designed to process sequential data and are vital components of systems that perform speech recognition \cite{2013graves+2}, machine translation \cite{2016wu+many}, handwritten text recognition \cite{2017capes+many}, and other tasks~\cite{2015schmidhuber}. 

An intuitively designed RNN is prone to gradient explosions or vanishing \cite{1994bengio+2} due to its recurrent nature. The impact of a~given input on future outputs of the RNN may vanish or explode with time. Specialized architectures with gates, namely Long Short-Term Memory (LSTM) networks \cite{1997hochreiter+1} and Gated Recurrent Unit (GRU) networks \cite{2014cho+6}, are designed to overcome this problem at the level of a~single neuron. While these networks are widely successful, they come with a cost --- their memory state undergoes only single-layer transformation from one time instant to another. 

Several recurrent architectures apply deep processing of their internal states \cite{2013pascanu+2,2015chung+3,2017zilly+3}. However, they are complex or challenging to train. 

This paper addresses the above shortcomings by introducing a neural module designed to prevent the previously mentioned gradient problems while allowing the state transformation to be modelled by an arbitrary feedforward neural network. We call this module Deep Memory Update (DMU). 
\footnote{We make the code available at \url{https://github.com/fuine/dmu}}. 
As a result, state transformation can easily be shaped in DMU. Additionally, the architecture is resistant to problems of gradient exploding/vanishing. Experimental results presented in the paper confirm that DMU
performs well in comparison to its state-of-the-art counterparts.

RNNs are often outperformed by feedforward networks with attention, especially by the transformer~\cite{2017vaswani+7}. However, the computational complexity of these  techniques excludes them from some applications \cite{2021jia+5,2020hansen+6}. It is also likely that some combination of attention and RNNs, such as R-Transformer \cite{2019wang+3}, ASRNN \cite{2021lin+3} and others \cite{2020liu+4}, will outperform both. Therefore, in this paper, we focus solely on RNNs. 

\section{Related work} 
\label{sec:related-work} 

Early RNNs \cite{1986jordan,1990elman,1987robinson+1,1988werbos} suffered from the problem of gradient vanishing/exploding, defined by \cite{1994bengio+2}: A~small change in the RNN's weights causes its future output's change that is vanishing or exploding in time. As a~result, the impact of RNN's weights on its performance is either close to zero or infinity. In either case, it is impossible to train such a~network. A~gradient norm clipping strategy proposed in \cite{2013pascanu+2} may mitigate this problem to some extent. \cite{arjovsky2016unitary} used orthogonal matrices of weights in shallow RNNs to stabilize the gradient successfully. 

The gradient vanishing/exploding problem was alleviated at a cell level with Long Short-Term Memory (LSTM) networks \cite{1997hochreiter+1}. A~neuron in such a~network is a~state machine with several so-called gates. The neuron generally preserves its state from one time to another but may also change it. The change depends on the dot product of the neuron inputs and its weights computed in its gates. LSTMs have been enhanced with batch normalization of a recurrent signal \cite{2017cooijmans+4}. 

\cite{2014cho+6} proposed an architecture based on neurons simpler than those in LSTMs, called Gated Recurrent Units (GRUs). Despite its simplicity, it generally preserved the favourable properties of LSTM. \cite{2018li+4} proposed a~unit whose state was only computed based on its previous state and the outputs of the preceding neural layer. Networks based on such units, Independently Recurrent Neural Networks (IndRNNs), tend to outperform LSTMs and GRUs. 

Capturing long-term dependencies in input sequences is a~crucial challenge that RNNs face. \cite{2017chang+9} proposed to increase the lag of recurrent connections in higher network layers geometrically. \cite{2018campos+4} introduced SkipRNN that learns to skip state updates and shorten the effective size of the computational graph. \cite{2018tallec+1} prove that RNNs operate via transformations of time, and the gates in LSTM and GRU networks are a~straightforward way to perform these transformations. 

LSTMs and GRUs are usually organized in several layers stacked on top of one another \cite{2013graves}. Input to each neuron within a~layer includes the previous states of all the neurons in the layer. This way, at each time instant, the network input undergoes a~deep transformation. However, the internal state of the network undergoes only a~shallow, single-layer transformation. 

Being able to apply an~arbitrary nonlinear, deep transformation to its internal state is a~valuable feature of a~recurrent neural network. \cite{2014pascanu+3} proposed to increase the recurrence depth by adding multiple nonlinear layers to the recurrent transition, resulting in Deep Transition RNNs (DT-RNNs) and Deep Transition RNNs with Skip connections (DT(S)-RNNs). Gradient propagation issues are exacerbated in these architectures due to long credit assignment paths. \cite{2015chung+3} added extra connections between all states across consecutive time steps in a stacked RNN, which also increases recurrence depth. However, their model requires additional connections with increasing depth, gives only a fraction of state cells access to the deepest layers, and faces gradient propagation issues along the longest paths.

\cite{2017zilly+3} introduced Recurrent Highway Networks (RHNs), which can be understood as LSTMs with specialized multilayer gates. These networks apply deep processing to their internal state while successfully coping with gradient vanishing/exploding. However, our proposed architecture requires only two state-processing gates as opposed to LSTM's three. Additionally, DMU allows for an arbitrary feedforward network to process the state.

A~number of concepts may facilitate the performance of RNNs. \cite{le2015simple} proposed a~scheme of initialization of weights in these networks. RNNs are usually trained with Stochastic Gradient Descent with gradient estimates computed with backpropagation through time. However, recent work of \cite{2021kag+1} on forward propagation through time calls this practice into question. An interesting alternative to gated recurrent neural networks is network simulators of continuous dynamical systems \cite{Kag2020RNNs,erichson2020lipschitz,chang2018antisymmetricrnn,NEURIPS2018_7bd28f15}.

\section{Method} 
\label{sec:method} 

In this section, we introduce the Deep Memory Update (DMU) module. It is a~neural module with memory designed to have the following properties: 
\begin{enumerate} 
\item 
Its memory state can undergo an arbitrary nonlinear transformation from one moment to another.
\item 
The module can easily preserve its memory state from one moment of time to another. 
\item 
Its learning is relatively fast and stable. 
\end{enumerate}

\subsection{General structure} 
\label{sec:gen stru} 

\begin{figure}
    \centering
    \includegraphics[width=\linewidth]{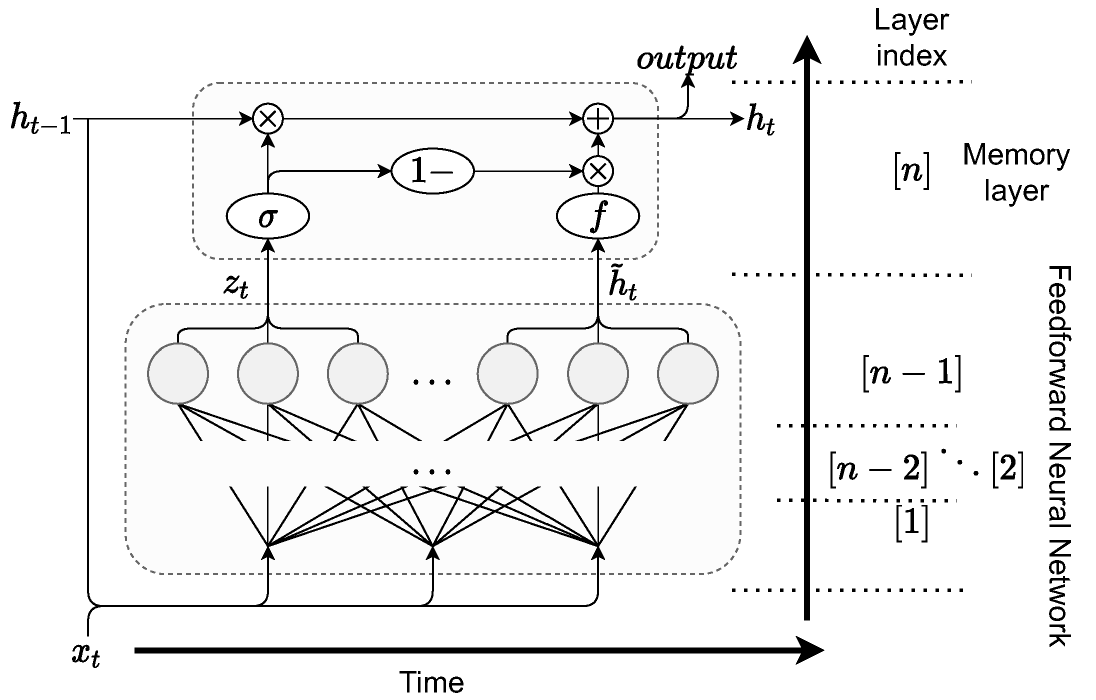}
    \caption{Structure of Deep Memory Update module. The module comprises the feedforward neural network, which can arbitrarily process the state and a memory layer. The output of the module is also its hidden state.} 
    \label{fig:DMU-structure}
\end{figure}

We present the structure of the Deep Memory Update (DMU) module in Fig.~\ref{fig:DMU-structure}. The module operates in discrete time $t=1,2,\dots$. At each time, the module is fed with the input $x_t \in \real^m$ and produces the vector $h_t \in \real^d$, which is both its~memory state and its~output. 

A~lagged memory state, $h_{t-1}$, together with an~input of the block, $x_t$, are fed to a~feedforward neural network, FNN. The network's output layer is linear with $2d$ neurons. It produces two vectors: $z_t \in \real^d$ determines to what extent the memory state should be preserved, and $\est h_t \in \real^d$ determines the direction in which the state should change. 

A pair of $i$-th elements of $z_t$ and $\est h_t$ are fed to a $i$-th memory cell. The new cell state is a~weighted, with $z_t$, average of the old state, $h_{t-1}$, and $\est h_t$. The memory state update takes the form  
\Beqs 
    \langle z_t, \est h_t \rangle & = \text{FNN}(h_{t-1},x_t) \label{zh=FNN} \\ 
    h_t & = h_{t-1} \circ \sigma(z_t) + f(\est h_t) \circ (\ones-\sigma(z_t)), \label{h:update} 
\Eeqs
where ``$\circ$'' denotes the elementwise product, $\ones$ is a~vector of ones, $\sigma$ is a~unipolar soft step function, e.g. the logistic sigmoid, 
\Beq \label{sigma:logistic} 
    \sigma_i(z) = \frac{e^{z_i}}{1+ e^{z_i}} 
    \; \text{for} \; z_i \in \real,   
\Eeq
and $f$ is an activation function, e.g. 
\Beq \label{f:tanh} 
    f_i(z) = \text{tanh}(z_i) 
    \; \text{for} \; z_i \in \real. 
\Eeq
Our proposed recurrent architecture is compared with GRU \cite{2014cho+6} in the supplementary material. 

Let us consider how the required properties of DMU are achieved. 
\begin{enumerate} 
\item 
Since a feedforward neural network with at least two dense layers is a~universal function approximator, the network state can undergo the arbitrary nonlinear transformation from one time moment to another.
\item 
The block preserves its memory state for large values of $z_t$. In particular, for $z_t = +\infty$ we have $h_t = h_{t-1}$. 
\item 
For efficient and stable training of the network, it is enough that the learning rate of the module is sufficiently lower than that of the rest of the network, as discussed in Section~\ref{sec:training}. 
\end{enumerate}

\subsection{Initialization} 

The FNN block should be a universal approximator. It can be a~multilayer perceptron with at least two layers, including a~linear output layer. This layer needs to be linear because its output should not be limited. It should be possible that $z_t\gg1$ which causes the memory state to be preserved, $h_t \cong h_{t-1}$. 

We recommend using the standard ways of initializing neural weight matrices in the FNN block, with one exception. Namely, upon weights' initialization, we recommend adding a~positive scalar to the biases of the neurons that produce $z_t$ values, e.g., 3. With positive elements of $z_t$, the memory state of the DMU module will be, by default, largely preserved from one moment $t$ to another. This addition is optional in most of the tasks, however if the network initialized in the standard way fails to converge, the positive bias usually helps.

We use Xavier initialization~\cite{glorot2010understanding} in all of the experiments. Additionally, in synthetic tasks, we use the positive bias with a value of 3.

\subsection{Training} 
\label{sec:training} 

Training of DMU may be based on gradient backpropagation through time and using the gradient with a~method of stochastic optimization such as Stochastic Gradient Descent or ADAM \cite{2014kingma+1}. These methods apply a~learning rate to each trained weight. In turn, the learning rate defines a~speed of optimization along derivatives with respect to this weight. Typically, the learning rates are equal for all weights. 

Let us consider DMU as a module in~a feedforward architecture. Its learning speed and stability can be noticeably improved by distinguishing a~module's learning rate and setting its value smaller than that of the rest of the architecture. The learning of recurrent modules is exposed to instability, which naturally limits its learning speed. Nevertheless, it does not need to limit the learning speed of the surrounding feedforward modules, which are less exposed to instability, and thus may learn faster.

In our experiments in Sec.~\ref{sec:experiments}, we combine $n$-layer DMU modules with $n'$-layer feedforward output subnetworks. For $\beta>0$ being a~learning rate for the output subnetwork we use a~learning rate of the DMU module, $\beta_\text{DMU}$, equal to 
\Beq \label{DMU:lr} 
    \beta_\text{DMU} = \frac\beta{2n}. 
\Eeq 
The deeper the DMU module, the lower its learning rate.  Additionally, when weight decay is used in the network training, its strength in the DMU module is reduced $2n$ times. 

\comment{
The gates in LSTM, GRU and RHN play the following role: The state of the network is generally preserved, and the gradient flows back through the time at a similar magnitude. The gates learn to identify parts of the state that need to be incrementally changed according to the current input and state. DMU does not introduce any novelty in this particular mechanism because it is also based on gates. The parts of the state that need an~increment and the increment alone can be identified accurately due to the DMU's gates, which can generally represent any nonlinear transformation. 
}

\subsection{Gradient propagation in DMU} 

\begin{figure}[h]
    \centering
    \includegraphics[width=\linewidth]{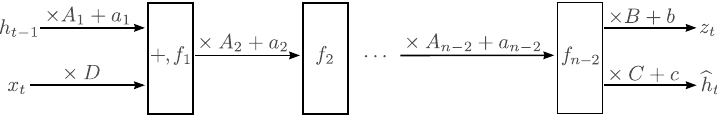}
    \caption{Structure of the feedforward module inside DMU. $A_i$, $B$, $C$ and $D$ denote weight matrices, $a_i$, $b$ and $c$ denote vectors of biases and $f_i$ denotes activation functions. } 
    \label{fig:FNN-structure}
\end{figure}

In order to analyze gradient propagation in DMU, we adopt the following further assumptions and notation: 
\begin{itemize} 
\item
The detailed structure of the FNN inside DMU is presented in Fig.~\ref{fig:FNN-structure}, with $A_i, B, C, D$ denoting weight matrices, $a_i, b, c$ denoting bias vectors and $f_i$ denoting activation functions. 
\item 
Activation functions in hidden layers of FNN are bipolar sigmoids with derivatives and absolute values covering the intervals $(0,1]$ and $(-1,1)$, respectively. 
\item 
$\sigma$ takes the form \eqref{sigma:logistic}. Therefore, $\sigma'(z)=\sigma(z)(1-\sigma(z))<1-\sigma(z)$. 
\item 
Vectors considered are in row form. 
\end{itemize} 
Note that $h_t$ are in fact weighted averages, over $i\geq0$, of $f(\est h_{t-i})$ never exceeding $(-1,1)$. Therefore, the elements of $h_t$ also never exceed $(-1,1)$. 

Let us analyze how the loss $L_{t'}$ resulting from the network output at time $t'$ propagates back to time $t-1<t'$. We have the following recursion: 
\Beqs 
    & \frac{\gd L_{t'}}{\gd h_{t-1}} = \frac{\gd L_{t'}}{\gd h_{t}} \frac{\gd h_t}{\gd h_{t-1}}  \notag\\ 
    & = \frac{\gd L_{t'}}{\gd h_{t}} \frac\gd{\gd h_{t-1}} \left(h_{t-1} \circ \sigma(z_t) + f(\est h_t) \circ (\ones -\sigma(z_t))\right)  \notag\\ 
    & = \frac{\gd L_{t'}}{\gd h_{t}} \bigg(\diag{\sigma(z_t)} + \r{h_{t-1}}\circ\frac{\gd z_t}{\gd h_{t-1}} \circ \r{\sigma'(z_t)} + \frac{\gd\est h_t}{\gd h_{t-1}} \circ \notag\\ 
    & \qquad \circ\r{f'(\est h_t)}\circ\r{\ones-\sigma(z_t)} - \r{f(\est h_t)}\circ\frac{\gd z_t}{\gd h_{t-1}} \circ\r{\sigma'(z_t)}\bigg) 
    \notag\\ 
    \begin{split} 
    & = \frac{\gd L_{t'}}{\gd h_{t}} \bigg(\diag{\sigma(z_t)} + \r{h_{t-1}-f(\est h_t)}\circ\frac{\gd z_t}{\gd h_{t-1}} \circ \r{\sigma'(z_t)}  \\ 
    & \qquad\qquad + \frac{\gd\est h_t}{\gd h_{t-1}}\circ\r{f'(\est h_t)}\circ\r{\ones-\sigma(z_t)}\bigg),
    \end{split} \label{first:component}
\Eeqs
where $\diag{v}$ denotes the diagonal matrix with the vector $v$ on its diagonal and $\r{v}$ denotes the matrix with the same vector $v^T$ in each column. 

By neglecting activation functions inside the FNN block, we reduce it to a~cascade of linear transformations and obtain the following approximations of the Jacobi matrices in \eqref{first:component}: 
\Beq \label{Jacobians} 
    \frac{\gd z_t}{\gd h_{t-1}} \cong B^T\left(\prod_{i=1}^{n-2} A_i\right)^T, \quad 
    \frac{\gd\est h_t}{\gd h_{t-1}} \cong 
    C^T\left(\prod_{i=1}^{n-2} A_i\right)^T. 
\Eeq
Considering that $\sigma(z_t)\in(0,1)$, $h_{t-1}-f(\est h_t) \in (-2,2)$, $\sigma'(z_t)\in(0,1-\sigma(z_t))$, $f'(\est h_t)\in(0,1)$, we obtain the following condition on non-increasing gradient: 
\begin{itemize} 
\item[$\star$] Eigenvalues of the matrices $B^T\left(\prod_{i=1}^{n-2} A_i\right)^T$ and $C^T\left(\prod_{i=1}^{n-2} A_i\right)^T$ remain in the intervals $(-1/2,1/2)$ and $(-1,1)$, respectively. 
\end{itemize} 
Essentially, that means that the components of the weight matrices in the FNN block should not be too large. 

When the above condition ($\star$) is satisfied, the gradient decreases when propagated back according to the first component of \eqref{first:component}, that is, by a factor of $\sigma(z_t)$. Intuitively, when the memory state $h_{t-1}$ is preserved to another time-step proportionally to $\sigma(z_t)$, the impact of this memory state on future performance is preserved likewise. 

\section{Experimental study} 
\label{sec:experiments} 

To evaluate the DMU architecture, we test it on three synthetic problems and three modern problems based on real-life data. The synthetic problems are taken from \cite{1997hochreiter+1}, and are noisy sequences, adding, and temporal order. The modern data-based problems are polyphonic music modelling \cite{boulangerlewandowski2012modeling}, natural language modelling \cite{2014zaremba+2}, and Spanish/German/Portuguese to English machine translation tasks \cite{2020tatoeba,2020manythings}.

We compare our DMU module using shallow architectures with ordinary recurrent neural networks (RNNs), GRU, LSTM, and RHN in the synthetic problems. We also compare DMU in its deep version with RHN in the data-based problems. To make the comparison fair, we embed a~recursive subnetwork within the same neural architecture. That subnetwork is a~layer or a~few layers of recurrent units or a~DMU module or RNH. Moreover, for each depth of RNNs, we compare different architectures of similar sizes measured by the number of weights.  

A reader may find details of our experimental setting, hyperparameters of architectures and their training in the supplementary material.

\subsection{Adding problem} 

The first task will be called “Adding”. It is taken from \cite[sec. 5.4]{1997hochreiter+1}. 

\begin{figure*}[!h]
    \centering
    \includegraphics[width=\textwidth]{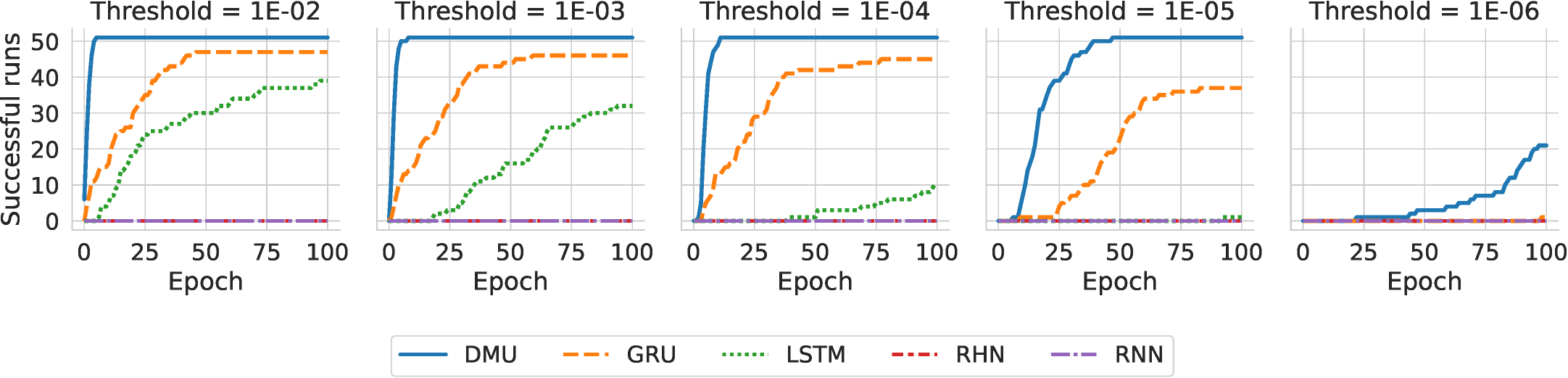}
    \caption{Adding: Results of 51 runs, five graphs for different loss thresholds, a~curve presents how many runs reach a~given loss threshold at a~given training epoch.}
    \label{fig:adding:res}
\end{figure*}

\paragraph{Results.} We present the results for the adding problem in \cref{fig:adding:res}. We conclude that DMU significantly outperforms all other modules, and GRU scores better than LSTM. RNN and RHN are not able to reach any threshold within 100 training epochs for any hyperparameters.

\subsection{Temporal order} 

The next task, referred to as ``TempOrd'', is taken from \cite[sec. 5.6, Task 6b]{1997hochreiter+1}. 

\begin{figure*}[!h]
    \centering
    \includegraphics[width=\textwidth]{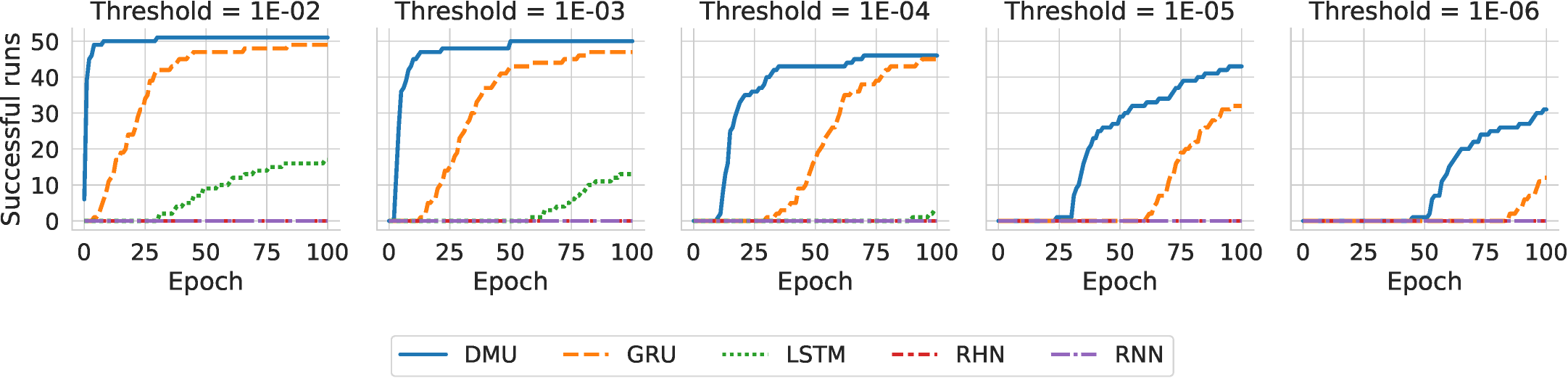}
    \caption{TempOrd: Results of 51 runs, five graphs for different loss thresholds.}
    \label{fig:tempord:res}
\end{figure*}

\paragraph{Results.} The results for the TempOrd task are depicted in \cref{fig:tempord:res}. We note that DMU has faster convergence than GRU and maintains similar results for high thresholds (up to $10^{-4}$). For lower thresholds, DMU outperforms GRU. LSTM reaches partial success on higher thresholds but fails for lower ones. RNN and RHN fail for all thresholds without a single successful 100 epoch run. 

\subsection{Noise-free and noisy sequences}

We call this task ``NoiseSeq''. It is taken from \cite[sec. 5.2]{1997hochreiter+1}. 

\begin{figure*}[!h]
    \centering
    \includegraphics[width=\textwidth]{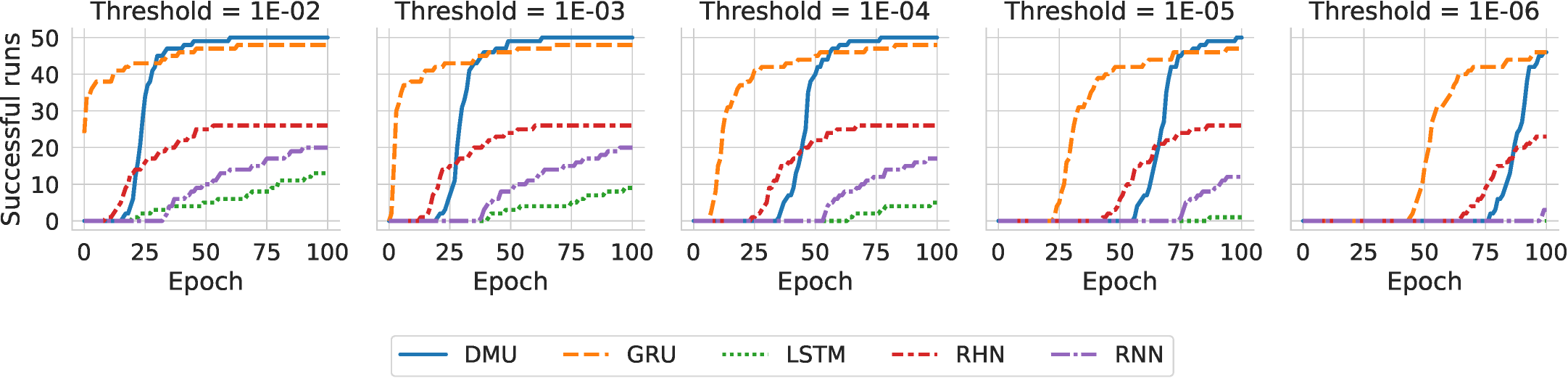}
    \caption{NoiseSeq: Results of 51 runs, five graphs for different loss thresholds.}
    \label{fig:noiseseq:res}
\end{figure*}

\paragraph{Results.} \Cref{fig:noiseseq:res} contains the results for the NoiseSeq task. We observe that GRU and DMU obtain similar results, in most cases reaching all the loss thresholds, with GRU training faster. RHN in about half of the cases does not reach any  threshold, and in the other half, it reaches all of them. RNN performs worse than RHN, and LSTM performs worse than RNN.

\subsection{Polyphonic music modelling}

In this subsection, we evaluate modules on the polyphonic music modelling task, referred to as ``PolyMusic'', based on the Nottingham music dataset~\cite{boulangerlewandowski2012modeling}.

\paragraph{Results.} The results of the polyphonic music modelling can be found in \cref{tab:polymusic:res}. 
In this problem, DMU outperforms RHN at 3 out of 4 depths with regard to test mean loss. 

\begin{table}[!h]
\renewrobustcmd{\bfseries}{\fontseries{b}\selectfont}
\sisetup{
         table-format=1.2,
         table-text-alignment=right,
         round-mode = places,
         detect-all = true,
         round-precision = 2,
detect-weight,mode=text
        }
\caption{PolyMusic: results --- loss. $N$ denotes the number of hidden layers.}
\label{tab:polymusic:res}
\centering
\begin{tabular}{l c S[table-format=3.2] S[table-format=3.2] S[table-format=2.2] S[table-format=3.2] S[table-format=3.2] S[table-format=2.2]}
\hline 
& & \multicolumn{3}{c}{train} & \multicolumn{3}{c}{test}\\
    {$N$} & {model} & {best} & {$\mu$} & {$\sigma$} & {best} & {$\mu$} & {$\sigma$}\\
\hline
\multirow{2}*{1} & RHN & 3.344 & 3.375 & 0.078 & \bfseries{3.552} & \bfseries{3.598} & \bfseries{0.037}\\
 & DMU & \bfseries 2.942 & \bfseries 2.959 & \bfseries 0.042 & 3.567 & 3.631 & 0.050\\
\hline
\multirow{2}*{2} & RHN & 3.390 & 3.414 & 0.098 & 3.553 & 3.607 & 0.063\\
 & DMU & \bfseries 3.022 & \bfseries 3.094 & \bfseries 0.067 & \bfseries 3.487 & \bfseries 3.551 & \bfseries 0.041\\
\hline
\multirow{2}*{5} & RHN & 3.443 & 3.682 & 0.163 & 3.734 & 3.851 & 0.106\\
 & DMU & \bfseries 3.215 & \bfseries 3.208 & \bfseries 0.064 & \bfseries 3.630 & \bfseries 3.685 & \bfseries 0.032\\
\hline
\multirow{2}*{10} & RHN & 3.701 & 3.927 & \bfseries 0.143 & \bfseries 3.903 & 4.075 & 0.118\\
 & DMU & \bfseries 3.523 & \bfseries 3.537 & 0.149 & 3.951 & \bfseries 4.044 & \bfseries 0.087\\
\hline
\end{tabular}
\end{table}

\subsection{Natural language modelling} 

The task called ``NatLang'' is based on the Penn Treebank corpus of English \cite{1993marcus+2}.

\paragraph{Results.} \cref{tab:natlang:res} shows the results.
DMU achieves consistently better results than RHN, often by a~large margin. 
Only for a~depth of 2 RHN performs slightly better than DMU with respect to the mean test perplexity.

\begin{table}[!h]

\renewrobustcmd{\bfseries}{\fontseries{b}\selectfont}
\sisetup{
         table-format=1.2,
         table-text-alignment=right,
         round-mode = places,
         detect-all = true,
         round-precision = 2,
detect-weight,mode=text
        }
\caption{NatLang: results --- perplexity (lower = better). $N$ is the depth of the network.}
\label{tab:natlang:res}
\centering
\begin{tabular}{l c S[table-format=3.2] S[table-format=3.2] S[table-format=2.2] S[table-format=3.2] S[table-format=3.2] S[table-format=2.2]}
\hline 
& & \multicolumn{3}{c}{train} & \multicolumn{3}{c}{test}\\
    {$N$} & {model} & {best} & {$\mu$} & {$\sigma$} & {best} & {$\mu$} & {$\sigma$}\\
\hline
\multirow{2}*{1} & RHN & 61.102 & 66.531 & 6.071 & 106.105 & 110.394 & 4.363\\
 & DMU & \bfseries 56.625 & \bfseries 59.422 & \bfseries 3.137 & \bfseries 105.354 & \bfseries 106.129 & \bfseries 0.475\\
\hline
\multirow{2}*{2} & RHN & \bfseries 62.895 & 66.317 & 4.232 & \bfseries 104.888 & \bfseries 109.833 & 4.040\\
 & DMU & 64.421 & \bfseries 64.861 & \bfseries 1.434 & 109.601 & 110.128 & \bfseries 0.465\\
\hline
\multirow{2}*{5} & RHN & \bfseries 82.325 & \bfseries 86.097 & 3.832 & 123.158 & 124.971 & 1.923\\
 & DMU & 92.395 & 94.063 & \bfseries 1.114 & \bfseries 117.924 & \bfseries 120.368 & \bfseries 1.456\\
\hline
\multirow{2}*{10} & RHN & \bfseries 85.972 & 149.430 & 86.869 & \bfseries 124.458 & 171.598 & 60.383\\
 & DMU & 118.199 & \bfseries 119.475 & \bfseries 1.526 & 130.054 & \bfseries 131.827 & \bfseries 1.440\\
\hline
\end{tabular}
\end{table}

\subsection{Machine translation} 

Next, we test the modules in the context of machine translation using recurrent architectures. The task is based on datasets of pairs of corresponding Spanish/Portuguese/German and English sentences \cite{2020tatoeba,2020manythings}. We will call experiments based on subsequent pairs ``Spa2Eng'', ``Por2Eng'', and ``Ger2Eng''. 

\paragraph{Results.} \Cref{tab:translation_acc:res} contains the results. DMU achieves a better perplexity score than RHN for all three language pairs at each depth of both networks except for Portuguese at depth 1. Additionally, both networks achieve the best results for a~depth of 1 or 2. Performance generally deteriorates with growing depth, significantly faster for RHN than for DMU.

\begin{table}[!h]
\renewrobustcmd{\bfseries}{\fontseries{b}\selectfont}
\sisetup{
         table-format=1.2,
         table-text-alignment=right,
         round-mode = places,
         detect-all = true,
         round-precision = 2,
detect-weight,mode=text
        }
\caption{Translation: results --- perplexity (lower = better). l. -- language pair.}
\label{tab:translation_acc:res}
\centering
\begin{tabular*}{\columnwidth}{c p{0.2cm} p{0.4cm} S[table-format=2.2] S[table-format=2.2] S[table-format=2.2] S[table-format=3.2] S[table-format=3.2] S[table-format=2.2]}
\toprule
& & & \multicolumn{3}{c}{train} & \multicolumn{3}{c}{test}\\
    l. & {$N$} & {model} & {best} & {$\mu$} & {$\sigma$} & {best} & {$\mu$} & {$\sigma$}\\
    
\midrule
\multirow{8}*{\rotatebox[origin=c]{90}{Spa2Eng}} & \multirow{2}*{1} & RHN & 8.257 & 7.720 & 0.285 & \bfseries 5.804 & 6.310 & 0.357\\
&  & DMU & \bfseries 7.128 & \bfseries 7.242 & \bfseries 0.171 & 5.886 & \bfseries 6.045 & \bfseries 0.137\\
\cmidrule{2-9}
& \multirow{2}*{2} & RHN & \bfseries 8.818 & \bfseries 8.498 & \bfseries 0.222 & \bfseries 6.532 & \bfseries 7.101 & 0.358\\
&  & DMU & 9.072 & 8.515 & 0.375 & 6.913 & 7.181 & \bfseries 0.316\\
\cmidrule{2-9}
& \multirow{2}*{5} & RHN & 12.379 & 24.678 & 12.479 & 12.337 & 47.599 & 41.679\\
&  & DMU & \bfseries 7.738 & \bfseries 7.909 & \bfseries 0.347 & \bfseries 7.503 & \bfseries 8.016 & \bfseries 0.345\\
\cmidrule{2-9}
& \multirow{2}*{10} & RHN & 58.744 & 58.730 & 0.924 & 110.535 & 141.766 & 56.012\\
&  & DMU & \bfseries 8.781 & \bfseries 8.837 & \bfseries 0.264 & \bfseries 7.988 & \bfseries 8.395 & \bfseries 0.234\\

\midrule

\multirow{8}*{\rotatebox[origin=c]{90}{Por2Eng}} & \multirow{2}*{1} & RHN & 3.949 & 3.977 & \bfseries 0.090 & 3.652 & 3.801 & 0.157\\
&  & DMU & \bfseries 3.908 & \bfseries 3.932 & 0.108 & \bfseries 3.541 & \bfseries 3.680 & \bfseries 0.104\\
\cmidrule{2-9}
& \multirow{2}*{2} & RHN & \bfseries 4.290 & \bfseries 4.328 & \bfseries 0.058 & \bfseries 3.669 & \bfseries 3.932 & 0.176\\
&  & DMU & 4.626 & 4.448 & 0.141 & 3.744 & 3.967 & \bfseries 0.137\\
\cmidrule{2-9}
& \multirow{2}*{5} & RHN & 6.250 & 7.500 & 0.833 & 6.553 & 7.616 & 0.849\\
&  & DMU & \bfseries 4.650 & \bfseries 4.740 & \bfseries 0.088 & \bfseries 4.560 & \bfseries 4.692 & \bfseries 0.096\\
\cmidrule{2-9}
& \multirow{2}*{10} & RHN & 48.578 & 48.348 & 0.374 & 79.098 & 99.863 & 23.798\\
&  & DMU & \bfseries 5.117 & \bfseries 5.287 & \bfseries 0.184 & \bfseries 4.921 & \bfseries 5.060 & \bfseries 0.115\\
 
\midrule

\multirow{8}*{\rotatebox[origin=c]{90}{Ger2Eng}} & \multirow{2}*{1} & RHN & 4.661 & 4.633 & 0.162 & 4.272 & 4.347 & \bfseries 0.064\\
&  & DMU & \bfseries 4.277 & \bfseries 4.504 & \bfseries 0.120 & \bfseries 4.104 & \bfseries 4.207 & 0.076\\
\cmidrule{2-9}
& \multirow{2}*{2} & RHN & \bfseries 5.260 & \bfseries 5.100 & 0.113 & \bfseries 4.410 & \bfseries 4.594 & \bfseries 0.103\\
&  & DMU & 5.377 & 5.283 & \bfseries 0.068 & 4.562 & 4.793 & 0.190\\
\cmidrule{2-9}
& \multirow{2}*{5} & RHN & 8.132 & 9.242 & 1.237 & 7.930 & 9.179 & 1.275\\
&  & DMU & \bfseries 5.326 & \bfseries 5.415 & \bfseries 0.121 & \bfseries 5.212 & \bfseries 5.331 & \bfseries 0.116\\
\cmidrule{2-9}
& \multirow{2}*{10} & RHN & 47.994 & 48.407 & 0.291 & 83.470 & 134.791 & 92.367\\
&  & DMU & \bfseries 5.912 & \bfseries 5.824 & \bfseries 0.231 & \bfseries 5.631 & \bfseries 5.738 & \bfseries 0.073\\

\bottomrule

\end{tabular*}
\end{table}

\subsection{Ordered and permuted MNIST}
Finally, we compare DMU to selected state-of-the-art modules on the pixel-by-pixel MNIST image classification problem~\cite{le2015simple}. Each image is represented as a flattened array of pixels, and the module processes it one after another. Such setup allows us to evaluate the internal state drift on long inputs, as each image contains 784 pixels. The task comes in two flavors - sequential, in which each image is flattened in a row-wise manner and permuted, in which we apply the same random permutation to each image after flattening. 

\paragraph{Results.} \Cref{tab:mnist:res} contains the results. 
\begin{table}[!b]
	\caption{Test accuracy on ordered and permuted pixel-by-pixel MNIST.}
	\label{tab:mnist:res}
	\centering
		\begin{tabular}{l c c c c}
			\toprule
			Name                  &  ordered & permuted  & N & \# params\\
			\midrule 
			
			LSTM baseline by~\cite{arjovsky2016unitary} & 97.3\% & 92.7\% &128 & $\approx$68K\\        

			MomentumLSTM~\cite{nguyen2020momentumrnn} & 99.1\% & 94.7\% & 256 & $\approx$270K\\ 			
			
			Unitary RNN~\cite{arjovsky2016unitary} & 95.1\% & 91.4\% &512 & $\approx$9K\\  		
			
			Full Capacity Unitary RNN~\cite{wisdom2016full} & 96.9\% & 94.1\% &512 & $\approx$270K\\  
			
			Soft orth. RNN~\cite{vorontsov2017orthogonality} & 94.1\% & 91.4\% &128 & $\approx$18K\\

			Kronecker RNN~\cite{jose2018kronecker} & 96.4\% & 94.5\% & 512 & $\approx$11K\\

			Antisymmteric RNN~\cite{chang2018antisymmetricrnn}  & 98.0\% & 95.8\% &128 & $\approx$10K\\
			
			Incremental RNN~\cite{Kag2020RNNs} & 98.1\% & 95.6\% & 128 & $\approx$4K/8K\\

			Exponential RNN \cite{lezcano2019cheap} & 98.4\% & 96.2\% & 360 & $\approx$69K\\
			
			Sequential NAIS-Net~\cite{NEURIPS2018_7bd28f15}     &    94.3\%     &  90.8\%         & 128              &  $\approx$18K  \\      
			Lipschitz RNN~\cite{erichson2020lipschitz} & \textbf{99.4\%} & \textbf{96.3\%} & 128 & $\approx$34K\\
			\midrule
            DMU (ours) & 98.5\% & 93.4\% & 96 & $\approx$20K\\
            DMU (ours) & 98.7\% & 93.4\% & 128 & $\approx$34K\\
			\bottomrule
	\end{tabular}
\end{table}

\subsection{Learning rate ablation} 

We verify how a reduction of a~DMU learning rate according to~\eqref{DMU:lr} impacts the performance of the neural architecture with this module. In this order, we register the performance of each architecture with approximately optimized, with a~grid search, learning rate. In one variant, the learning rate is constant for the whole architecture. In the other, the learning rate of the DMU module and the learning rate for the rest are bound with~\eqref{DMU:lr}. Numerical results of this ablation are presented in Tables.~\ref{tab:polymusic:ablation}--\ref{tab:translation_acc:ablation}. Note that this ablation does not make sense for the analyzed synthetic problems because the recurrent module is the entire architecture in these cases. The results confirm that efficiency benefits from reducing the learning rate of the DMU module.

\begin{table}[]
\renewrobustcmd{\bfseries}{\fontseries{b}\selectfont}
\sisetup{
         table-format=1.2,
         table-text-alignment=right,
         round-mode = places,
         detect-all = true,
         round-precision = 2,
detect-weight,mode=text
        }
\caption{PolyMusic: varied learning rate ablation results --- loss. DMU-C --- DMU with an equal learning rate for all modules.}
\label{tab:polymusic:ablation}
\centering
\begin{tabular}{l c S[table-format=3.2] S[table-format=3.2] S[table-format=2.2] S[table-format=3.2] S[table-format=3.2] S[table-format=2.2]}
\hline 
& & \multicolumn{3}{c}{train} & \multicolumn{3}{c}{test}\\
    {$N$} & {model} & {best} & {$\mu$} & {$\sigma$} & {best} & {$\mu$} & {$\sigma$}\\
\hline
\multirow{2}*{1} & DMU-C & \bfseries 2.717 & \bfseries 2.813 & 0.082 & \bfseries 3.342 & \bfseries 3.382 & \bfseries 0.035\\
 & DMU & 2.942 & 2.959 & \bfseries 0.042 & 3.567 & 3.631 & 0.050\\
\hline
\multirow{2}*{2} & DMU-C & \bfseries 2.991 & \bfseries 3.034 & 0.197 & \bfseries 3.430 & \bfseries 3.486 & 0.041\\
 & DMU & 3.022 & 3.094 & \bfseries 0.067 & 3.487 & 3.551 & \bfseries 0.041\\
\hline
\multirow{2}*{5} & DMU-C & 3.251 & 3.308 & 0.236 & 3.902 & 3.993 & 0.114\\
 & DMU & \bfseries 3.215 & \bfseries 3.208 & \bfseries 0.064 & \bfseries 3.630 & \bfseries 3.685 & \bfseries 0.032\\
\hline
\multirow{2}*{10} & DMU-C & 3.752 & nan & nan & 4.262 & 5.032 & 0.632\\
 & DMU & \bfseries 3.523 & \bfseries 3.537 & \bfseries 0.149 & \bfseries 3.951 & \bfseries 4.044 & \bfseries 0.087\\
\hline
\end{tabular}
\end{table}

\begin{table}[]
\renewrobustcmd{\bfseries}{\fontseries{b}\selectfont}
\sisetup{
         table-format=1.2,
         table-text-alignment=right,
         round-mode = places,
         detect-all = true,
         round-precision = 2,
detect-weight,mode=text
        }
\caption{NatLang: varied learning rate ablation results --- perplexity. DMU-C --- DMU with an equal learning rate for all modules.}
\label{tab:natlang:ablation}
\centering
\begin{tabular}{l c S[table-format=3.2] S[table-format=3.2] S[table-format=2.2] S[table-format=3.2] S[table-format=3.2] S[table-format=2.2]}
\hline 
& & \multicolumn{3}{c}{train} & \multicolumn{3}{c}{test}\\
    {$N$} & {model} & {best} & {$\mu$} & {$\sigma$} & {best} & {$\mu$} & {$\sigma$}\\
\hline
\multirow{2}*{1} & DMU-C & 68.988 & 71.116 & \bfseries 1.779 & 109.230 & 111.195 & 1.424\\
 & DMU & \bfseries 56.625 & \bfseries 59.422 & 3.137 & \bfseries 105.354 & \bfseries 106.129 & \bfseries 0.475\\
\hline
\multirow{2}*{2} & DMU-C & 81.311 & 81.014 & 1.691 & 117.555 & 118.166 & 0.647\\
 & DMU & \bfseries 64.421 & \bfseries 64.861 & \bfseries 1.434 & \bfseries 109.601 & \bfseries 110.128 & \bfseries 0.465\\
\hline
\multirow{2}*{5} & DMU-C & 108.896 & 579.677 & 235.391 & 138.133 & 542.204 & 202.036\\
 & DMU & \bfseries 92.395 & \bfseries 94.063 & \bfseries 1.114 & \bfseries 117.924 & \bfseries 120.368 & \bfseries 1.456\\
\hline
\multirow{2}*{10} & DMU-C & 696.302 & 697.572 & \bfseries 0.688 & 642.178 & 642.914 & \bfseries 0.436\\
 & DMU & \bfseries 118.199 & \bfseries 119.475 & 1.526 & \bfseries 130.054 & \bfseries 131.827 & 1.440\\
\hline
\end{tabular}
\end{table}

\begin{table}[!ht]
\renewrobustcmd{\bfseries}{\fontseries{b}\selectfont}
\sisetup{
         table-format=1.2,
         table-text-alignment=right,
         round-mode = places,
         detect-all = true,
         round-precision = 2,
detect-weight,mode=text
        }
\caption{Translation: varied learning rate ablation results --- accuracy. DMU-C --- DMU with an equal learning rate for all modules.}
\label{tab:translation_acc:ablation}
\centering
\begin{tabular}{c c c S[table-format=1.2] S[table-format=1.2] S[table-format=1.2] S[table-format=1.2] S[table-format=1.2] S[table-format=1.2]}
\toprule
& & & \multicolumn{3}{c}{train} & \multicolumn{3}{c}{test}\\
    l. & {$N$} & {model} & {best} & {$\mu$} & {$\sigma$} & {best} & {$\mu$} & {$\sigma$}\\
\midrule
\multirow{10}*{\rotatebox[origin=c]{90}{Spa2Eng}} & \multirow{2}*{1} & DMU-C & 0.910 & 0.851 & 0.140 & \bfseries 0.698 & 0.671 & 0.044\\
&  & DMU & \bfseries 0.930 & \bfseries 0.927 & \bfseries 0.003 & 0.697 & \bfseries 0.692 & \bfseries 0.005\\
\cmidrule{2-9}
& \multirow{2}*{2} & DMU-C & 0.844 & 0.762 & 0.196 & 0.660 & 0.600 & 0.106\\
&  & DMU & \bfseries 0.938 & \bfseries 0.937 & \bfseries 0.009 & \bfseries 0.691 & \bfseries 0.686 & \bfseries 0.005\\
\cmidrule{2-9}
& \multirow{2}*{5} & DMU-C & 0.710 & 0.660 & 0.055 & 0.617 & 0.586 & 0.033\\
&  & DMU & \bfseries 0.837 & \bfseries 0.829 & \bfseries 0.007 & \bfseries 0.661 & \bfseries 0.653 & \bfseries 0.005\\
\cmidrule{2-9}
& \multirow{2}*{10} & DMU-C & 0.336 & 0.299 & 0.019 & 0.352 & 0.305 & 0.024\\
&  & DMU & \bfseries 0.760 & \bfseries 0.749 & \bfseries 0.009 & \bfseries 0.629 & \bfseries 0.625 & \bfseries 0.003\\

\midrule

\multirow{10}*{\rotatebox[origin=c]{90}{Por2Eng}} & \multirow{2}*{1} & DMU-C & \bfseries 0.941 & 0.929 & 0.019 & \bfseries 0.777 & 0.771 & 0.005\\
&  & DMU & 0.926 & \bfseries 0.941 & \bfseries 0.008 & 0.776 & \bfseries 0.773 & \bfseries 0.003\\
\cmidrule{2-9}
& \multirow{2}*{2} & DMU-C & 0.937 & 0.925 & 0.021 & 0.754 & 0.745 & 0.013\\
&  & DMU & \bfseries 0.955 & \bfseries 0.952 & \bfseries 0.006 & \bfseries 0.775 & \bfseries 0.768 & \bfseries 0.004\\
\cmidrule{2-9}
& \multirow{2}*{5} & DMU-C & 0.766 & 0.625 & 0.203 & 0.699 & 0.590 & 0.164\\
&  & DMU & \bfseries 0.861 & \bfseries 0.861 & \bfseries 0.004 & \bfseries 0.731 & \bfseries 0.725 & \bfseries 0.005\\
\cmidrule{2-9}
& \multirow{2}*{10} & DMU-C & 0.315 & 0.316 & \bfseries 0.001 & 0.322 & 0.320 & \bfseries 0.002\\
&  & DMU & \bfseries 0.807 & \bfseries 0.803 & 0.012 & \bfseries 0.709 & \bfseries 0.706 & 0.003\\
 
\midrule

\multirow{10}*{\rotatebox[origin=c]{90}{Ger2Eng}} & \multirow{2}*{1} & DMU-C & 0.923 & 0.915 & 0.017 & 0.749 & 0.748 & \bfseries 0.001\\
&  & DMU & \bfseries 0.924 & \bfseries 0.925 & \bfseries 0.006 & \bfseries 0.753 & \bfseries 0.749 & 0.003\\
\cmidrule{2-9}
& \multirow{2}*{2} & DMU-C & 0.884 & 0.898 & 0.009 & 0.730 & 0.722 & 0.005\\
&  & DMU & \bfseries 0.935 & \bfseries 0.938 & \bfseries 0.005 & \bfseries 0.749 & \bfseries 0.743 & \bfseries 0.004\\
\cmidrule{2-9}
& \multirow{2}*{5} & DMU-C & 0.724 & 0.702 & 0.019 & 0.662 & 0.647 & 0.014\\
&  & DMU & \bfseries 0.845 & \bfseries 0.843 & \bfseries 0.010 & \bfseries 0.711 & \bfseries 0.703 & \bfseries 0.006\\
\cmidrule{2-9}
& \multirow{2}*{10} & DMU-C & 0.443 & 0.336 & 0.055 & 0.446 & 0.344 & 0.053\\
&  & DMU & \bfseries 0.797 & \bfseries 0.782 & \bfseries 0.011 & \bfseries 0.684 & \bfseries 0.679 & \bfseries 0.004\\

\bottomrule

\end{tabular}
\end{table}

\section{Discussion}
\label{sec:discussion} 

Since the seminal paper of \cite{1997hochreiter+1} the development of recurrent neural networks has been stimulated by the need to avoid gradient exploding or vanishing in backpropagation through time. Indeed, these phenomena are likely to occur in neural networks with feedback loops. In LSTM and GRU architectures, they were eliminated at the cell level. 

The DMU neural module introduced in this paper is based on memory cells whose state is updated with the weighted average of their previous content and new values proposed for them. Both the weights and the new proposed values come from a~feedforward subnetwork whose inputs include the previous state of the memory cells. Architectures based on the DMU module compete with and often outperform those based on LSTM or GRU. The gradient vanishing/exploding problem is solved in DMU at the module level.

In some applications, deep transformation of the network state is necessary. However, then the effective length of the gradient path increases, which may destabilize training. RHN successfully coped with this problem at the expense of the complexity of its architecture. DMU applies a~typical feedforward block of any depth for state transformation. Training stability is ensured by appropriately reducing the learning rate of the DMU module. As a result, DMU performed better than RHN of the same depth in all three analyzed data-based problems with a~handful of exceptions.

Interestingly, contrary to \cite{2017zilly+3} we note that depth-scaling of the model did not yield better results. We speculate that it can be explained by the lack of regularization other than weight decay. This was a deliberate choice to compare RHN and DMU modules without any unnecessary architectural additions.

In the future, we want to further investigate DMU's fast convergence rate on synthetic tasks. A greater understanding of the model's behaviour could help us improve the architecture and 
provide additional insight into the state drift problem of RNNs in general.

\section{Conclusions} 
\label{sec:conclusions} 

In this paper, we propose DMU --- a~recurrent neural module that can perform an arbitrary nonlinear transformation of its memory state. Three experiments with synthetic data (Adding, Temporal order, Noisy sequence) presented here compare neural architectures based on DMU with those based on RNN, LSTM, and GRU. DMU yields the best results in two of them while having results comparable to the best module in the third one. Three experiments with real-life data (Polyphonic music, Natural language modelling, Machine translation) compare neural architectures based on DMU with those based on Recurrent Highway Networks of the same depth. The architecture based on DMU outperformed RHN in 15 out of 20 analyzed mean test score cases while staying competitive in the other five cases. 

\section*{Acknowledgments}

The project was funded by POB Research Centre for
Artificial Intelligence and Robotics of Warsaw  University of Technology within the Excellence Initiative Program -- Research University (ID-UB). We gratefully acknowledge the contribution of Aleksander Zamojski, Lidia Wojciechowska and Monika Berlińska to the code of DMU. 

%\printbibliography
%\printbibitembibliography
%\bibliography{references}

\begin{appendices}

\newcommand{\wek}[1]{
	{\bf #1} 
}
\newcommand{\mat}[1]{
	{\bf #1} 
}
\renewcommand{\theequation}{A.\arabic{equation}}
\setcounter{equation}{0}

\section{Comparision of DMU and GRU} 

In the notation applied in this paper operation of a~GRU \cite{2014cho+6} layer can be expressed as  
\begin{align*} 
    r_t & = W_r x_t + U_r h_{t-1} + b_r \\ 
    \est h_t & = W_h x_t + U_h (\sigma(r_t) \circ h_{t-1}) + b_h \\ 
    z_t & = W_z x_t + U_z h_{t-1} + b_z \\ 
    h_t & = h_{t-1} \circ \sigma(z_t) + f(\est h_t) \circ ({\bf 1} - \sigma(z_t))
\end{align*} 
where $W_r, U_r, W_h, U_h, W_z, U_z$ and $b_r, b_h, b_z$ are matrices and vectors of weights. The operation of DMU is presented in eqs. (1) and (2). In the most straightforward configuration, this network is a~layer of linear units. Then 
\begin{equation} 
    \begin{split} 
    \est h_t & = W_h x_t + U_h h_{t-1} + b_h  \\ 
    z_t & = W_x x_t + U_x h_{t-1} + b_x \\ 
    h_t & = h_{t-1} \circ \sigma(z_t) + f(\est h_t) \circ ({\bf 1} - \sigma(z_t))
    \end{split} \label{DMU:basic}
\end{equation} 
Therefore, DMU is simpler in this basic configuration, thus having fewer weights per memory cell than a layer of GRUs, as it does not have the reset gate. In the general configuration, DMU can apply an arbitrary nonlinear transformation to its state, which GRU is unable to do. In practice, GRU layers are often stacked on one another which improves its performance on tasks that require complex nonlinear transformation of state. However, the state of the stacked GRU layers still can not be arbitrarily transformed in a single time instant since parts of this state are transformed within single layers. 

LSTM \cite{1997hochreiter+1} and RHN \cite{2017zilly+3} are based on different, much more complex equations with even more weights. LSTM has twice more weights per memory cell than DMU has in the basic configuration. 

\section{Experiments} 

\subsection{Architectures}

We present architectures for each problem in \cref{tab:architectures} and \cref{tab:architectures_dmu_rhn}. Corresponding hyperparameters can be found in Table~\ref{tab:architectures}. The recurrent subnetwork is characterized by the number of units in subsequent layers. For example, a~GRU subnetwork with two layers of 10 and 20 neurons will be briefly denoted by $(10,20)$. A~DMU block with two FNN layers of 10 and 20 neurons will be denoted by $(10,20,10)$ to account for the layer of memory cells within the block. In the data-based problems, we evaluate each module at varying depths. In all cases, the compared architectures have matching numbers of trained parameters. Hyperparameters for the models were selected based on the random and grid searches and then fine-tuned manually. The metric used to evaluate the hyperparameters was calculated on the validation subset in each case.

\begin{table} 
\caption{Architectures used the for the comparison of different neural modules in synthetic experiments.$^1$Recurrent block.}
\label{tab:architectures}
\centering
\begin{tabular}{l l l l l l l} 
\hline 
experiment &  & RNN & LSTM & GRU & RHN & DMU\\ 
\hline 
\multirow{2}*{NoiseSeq} & rc. blk$^1$ & (5, 5) & (2, 2) & (2, 3) & ((3, 3)) & ((5, 4))\\  
& weights no. & 595 & 880 & 687 & 672 & 573\\
\hline 
\multirow{2}*{Adding} & rc. blk$^1$ & (5, 5) & (2, 2) & (3, 2) & ((4, 3)) & ((5, 5))\\ 
& weights no. & 111 & 99 & 108 & 136 & 106\\ 
\hline 
\multirow{2}*{TempOrd} & rc. blk$^1$ & (6, 6) & (2, 3) & (2, 4) & ((4, 3)) & ((5, 6)) \\ 
& weights no. & 236 & 212 & 208 & 224 & 203\\
\hline
\end{tabular} 
\end{table}

\begin{table} 
\caption{Architectures used for the comparison of RHN and DMU. We report the number of neurons in feedforward layers. The last layer of the DMU's FNN on the Translation task always has 200 neurons. For the Translation task, weights' numbers are provided for Spa2Eng, Ger2Eng, and Por2Eng, respectively.}
\label{tab:architectures_dmu_rhn}
\centering
\begin{tabular}{l l l l l l} 
\hline 
experiment & depth & RHN & DMU & weights no.\\ 
\hline 
\multirow{4}*{PolyMusic} & 1 & 100 & 100 & 46.7K\\
& 2 & 100 & 122 & 66.9K\\
& 5 & 100 & 131 & 127K\\
& 10 & 100 & 136 & 228K\\
\hline 
\multirow{4}*{NatLang} & 1 & 100 & 100 & 1.7M\\
& 2 & 100 & 122 & 1.7M\\
& 5 & 100 & 131 & 1.8M\\
& 10 & 100 & 136 & 1.9M\\
\hline
\multirow{4}*{Translation} & 1 & 200 & 200 & 27.8M/36.6M/24.5M\\
& 2 & 200 & 340 & 28.0M/36.8M/24.7M\\
& 5 & 200 & 300 & 28.4M/37.3M/25.2M\\
& 10 & 200 & 300 & 29.2M/38.1M/26.0M\\
\hline
\end{tabular} 
\end{table}

\subsection{Training} 

The data is split into training, validation, and testing set. On synthetic problems, training continues until the loss reaches a specified threshold ($10^{-6}$) on the validation set or the training budget is depleted. The error is then registered on the testing set and presented here. We follow a similar procedure for real-life problems, except the training process is stopped once the optimizer reaches the final epoch.
All metrics are calculated using the model from the epoch with the best metric score on the validation set.

We run the experiment five times for each modern task/model/depth combination and aggregate the results. Standard result aggregation, such as averaging loss over time, would not be interpretable in the synthetic tasks since training is often unstable in these experiments. Therefore, the results for each synthetic problem are presented for multiple thresholds of the loss value. We plot the number of experiment runs that have reached the threshold in or before the specific epoch for each threshold. These thresholds allow us to assess how fast and how likely the module converges to a specific loss value. Thus, we can gain an insight into the quality of the module. Faster attainment of a specific threshold and convergence to lower thresholds are both desirable for the algorithm.

%The quicker the algorithm reaches a particular threshold, the better. Convergence to lower thresholds is also preferable.

Hyperparameters used for each experiment/neural module are presented in \cref{tab:hyperparameters_synth} and \cref{tab:hyperparameters}. We use ADAM optimizer to train all architectures.

\begin{table}[h] 
\caption{Hyperparameters used for synthetic tasks.}
\label{tab:hyperparameters_synth}
\centering
\begin{tabular}{l l l l l l l l} 
\toprule 
experiment & hyperpameter & RNN & LSTM & GRU & RHN & DMU\\ 
\midrule
\multirow{4}*{NoiseSeq} & learning rate & 0.01 & 0.002 & 0.05 & 0.05 & 0.02 \\ 
& seq. per epoch & 200 & 200 & 200 & 200 & 200\\
& min seq. length & 100 & 100 & 100 & 100 & 100\\
& max epochs & 100 & 100 & 100 & 100 & 100\\
\midrule 
\multirow{4}*{Adding} & learning rate & 0.01 & 0.001 & 0.05 & 0.02 & 0.02 \\ 
& seq. per epoch & 200 & 200 & 200 & 200 & 200\\
& min seq. length & 100 & 100 & 100 & 100 & 100\\
& max epochs & 100 & 100 & 100 & 100 & 100\\
\midrule 
\multirow{4}*{TempOrd} & learning rate & 0.01 & 0.005 & 0.02 & 0.02 & 0.05 \\ 
& seq. per epoch & 200 & 200 & 200 & 200 & 200\\
& min seq. length & 100 & 100 & 100 & 100 & 100\\
& max epochs & 100 & 100 & 100 & 100 & 100\\
\bottomrule 
\end{tabular} 
\end{table}

\begin{table}[] 
\caption{Hyperparameters used for each experiment and each neural module.}
\label{tab:hyperparameters}
\centering
\begin{tabular}{l l l l l} 
\toprule 
experiment & depth & hyperpameter & RHN & DMU\\ 
\midrule
\multirow{13}*{PolyMusic} & \multirow{1}*{all} & max epochs & 500 & 500\\
\cmidrule{2-5}
& \multirow{3}*{1} & learning rate & 0.005 & 0.005 \\ 
& & weight decay & 0.001 & 0.0001 \\ 
& & scheduler gamma & 1.0 & 1.0 \\
\cmidrule{2-5}
& \multirow{3}*{2} & learning rate & 0.005 & 0.005 \\ 
& & weight decay & 0.001 & 0.0001 \\ 
& & scheduler gamma & 1.0 & 1.0 \\
\cmidrule{2-5}
& \multirow{3}*{5} & learning rate & 0.005 & 0.005 \\ 
& & weight decay & 0.001 & 0.0001 \\ 
& & scheduler gamma & 1.0 & 1.0 \\
\cmidrule{2-5}
& \multirow{3}*{10} & learning rate & 0.005 & 0.002\\ 
& & weight decay & 0.001 & 0.0001\\ 
& & scheduler gamma & 1.0 & 1.0 \\
\cmidrule{1-5} 
\multirow{13}*{NatLang} & \multirow{1}*{all} & max epochs & 40 & 40\\
\cmidrule{2-5}
& \multirow{3}*{1} & learning rate & 0.02 & 0.02\\ 
& & weight decay & 0.0001 & 0.0001\\ 
& & scheduler gamma & 0.9 &  0.9 \\
\cmidrule{2-5}
& \multirow{3}*{2} & learning rate & 0.02 & 0.02 \\ 
& & weight decay & 0.0001 & 0.0001 \\ 
& & scheduler gamma & 0.9 & 0.9\\
\cmidrule{2-5}
& \multirow{3}*{5} & learning rate & 0.02 & 0.01 \\ 
& & weight decay & 0.0001 & 0.0001 \\ 
& & scheduler gamma & 0.9 & 0.98 \\
\cmidrule{2-5}
& \multirow{3}*{10} & learning rate & 0.02 & 0.02\\ 
& & weight decay & 0.0001 & 0.0001\\ 
& & scheduler gamma & 0.9 & 0.98 \\
\cmidrule{1-5}
\multirow{14}*{\makecell*{Spa2Eng/\\Por2Eng/\\Deu2Eng}} & \multirow{2}*{all} & teacher forcing ratio & 1.0 & 1.0 \\
& & max epochs & 50 & 50\\
\cmidrule{2-5}
& \multirow{3}*{1} & learning rate & 0.01 & 0.005 \\ 
& & weight decay & 0.0001 & 0.0001 \\ 
& & scheduler gamma & 0.9 & 0.9 \\
\cmidrule{2-5}
& \multirow{3}*{2} & learning rate & 0.01 & 0.01 \\ 
& & weight decay & 0.0001 & 0.0001\\ 
& & scheduler gamma & 0.9 & 0.9 \\
\cmidrule{2-5}
& \multirow{3}*{5} & learning rate & 0.01 & 0.003\\ 
& & weight decay & 0.0001 & 0.0001\\ 
& & scheduler gamma & 0.9 & 1.0\\
\cmidrule{2-5}
& \multirow{3}*{10} & learning rate & 0.01 & 0.003\\ 
& & weight decay & 0.0001 & 0.0001\\ 
& & scheduler gamma & 0.9 &  1.0\\
\bottomrule
\end{tabular} 
\end{table}

\subsection{Hardware}
Our experiments have been performed on a PC equipped with AMD\texttrademark Ryzen 1920X, 64GB RAM, 4×NVidia\texttrademark RTX 2070 Super.

\subsection{Testing strategy}
To evaluate synthetic tasks, we run an experiment for each module 51 times and aggregate the results. On real-life data tasks, we aggregate results over five runs for each recurrent module. We report metrics obtained in the \texttt{best} runs. These runs are selected based solely on their performance on the test set. Therefore, in some cases, metrics reported in the \texttt{best} column for the training dataset are worse than those in the \texttt{mean} column.

\subsection{Adding problem}

In this problem, the network is fed with two-dimensional vectors $[a,b]$, where $a$ is randomly chosen from the interval $[-1,1]$, and $b\in\{-1,0,1\}$ is a~marker: $-1$ denotes the first and last element of the sequence, there are two pairs marked by $1$, the rest are marked by $0$. The task of the network is to output the sum of $a$-s accompanied by $b$-s equal to $1$ at the end of the sequence. 
Each network analyzed is composed of a~recurrent block and a~layer with softmax activation.

\subsection{Temporal order} 

This task evaluates network's ability to model temporal ordering of data. The input and the output are both 8-dimensional. They represent one of 8 symbols by one-hot encoding. The input symbols are: $E$ (start), $B$ (end), $X$ or $Y$. $X$ or $Y$ occur at time $t_1$, $t_2$, $t_3$. In all three of these occurrences the choice of $X$ or $Y$ is random, the rest of a sequence is filled with symbols $a,b,c,d$ also selected at random. Sequence length is chosen randomly between 100 and 110. $t_1$, $t_2$, $t_3$ are selected randomly for each sequence, respectively between 10-20, 33-43 and 66-76. The output desired at the end of a~sequence is either $Q,R,S,U,V,A,B,C$, depending on the combination of symbols that has occurred at times $t_1$, $t_2$ and $t_3$. 
Each network analyzed is composed of a~recurrent block and a~layer with softmax activation.

\subsection{NoiseSeq} 

We use noisy sequences to test the modules on the long time lag problems.
The network is fed with symbols one-hot encoded in $n$-dimensional vectors. An input sequence is, with equal probability 0.5, either $(x,a_1, \dots,a_{n-2})$ or $(y,a_1,\dots,a_{n-2})$, where $x,y,a_1, \dots, a_{n-1}$ are selected on random prior to an experiment. The task of the network is to output the first symbol in the input sequence when at $n-1$-st step. 
Each analyzed neural network is composed of a~recurrent block and a~layer with softmax activation.

\subsection{PolyMusic} 

Inputs and outputs are 88-dimensional. They represent the binary encoding of possible piano-rolls at a current timestep (in MIDI note numbers, between 21 and 108 inclusive). Sequences vary in length. The task of the model is to predict the next time step in the sequence (i.e., output at time $t$ is equal to input at time $t+1$). The loss function is a negative log-likelihood averaged over all time steps in the dataset/batch.
The neural network is composed of a~recurrent block and a~layer with the sigmoid activation.

\subsection{NatLang} 

Inputs and outputs are single number representations of the most frequent words in English and special tokens such as ``unknown'' or ``end of sequence''. Sequences include 100 words. The goal of the network is to predict another word within the current sequence. The loss function is perplexity (categorical cross-entropy exponent). See \cite{2014zaremba+2} for details. 
The whole neural network comprises a~recurrent block, followed by a 100-neurons dense layer and an output layer with the softmax activation. For this experiment, the input word embedding is set to a small size (64) on purpose to limit overfitting.

\subsection{Machine Translation} 

We use tokens representing words, punctuation marks, \textit{sentence start}, and \textit{sentence end} in all languages. Each token is encoded as a single, unique number. The goal is to translate Spanish/Portuguese/German sentences into English ones using a~system with encoder-decoder architecture \cite{2014cho+3,2014cho+6,2014sutskever+2}. 
A~whole translator has encoder-decoder architecture. An~encoder is a~recurrent block. A~decoder is composed of a~recurrent block and a~layer with the softmax activation. Additionally, we use input and output embeddings of size 650.

\end{appendices}

\end{document}